\documentclass[letterpaper, 10 pt, conference]{ieeeconf}  % Comment this line out if you need a4paper

\IEEEoverridecommandlockouts                              % This command is only needed if 

\usepackage{graphicx} % for pdf, bitmapped graphics files
\usepackage{placeins}
\usepackage{float}
\usepackage{amsmath} % assumes amsmath package installed
\usepackage{amssymb}  % assumes amsmath package installed
\usepackage[table]{xcolor}
\usepackage{booktabs}
\usepackage{multirow}
\usepackage{array}
\usepackage{arydshln}

\makeatletter
\def\@IEEEtablecaptionsepspace{\vskip 2.5pt}
\let\ICRAoriginalmakecaption\@makecaption
\long\def\@makecaption#1#2{%
  \ifx\@captype\@IEEEtablestring
    \vspace*{6pt}% Top-margin clearance; table typography is unchanged.
    \setbox\@tempboxa\hbox{\small #1: #2}%
    \ifdim\wd\@tempboxa>\hsize
      \parbox[t]{\hsize}{\small\noindent #1: #2}%
    \else
      \hbox to\hsize{\small\hfil\box\@tempboxa\hfil}%
    \fi
    \@IEEEtablecaptionsepspace
  \else
    \ICRAoriginalmakecaption{#1}{#2}%
  \fi}
\makeatother

\definecolor{bestgreen}{RGB}{172,220,164}
\definecolor{secondlime}{RGB}{226,236,170}
\definecolor{thirdyellow}{RGB}{248,220,160}
\newcommand{\calib}[1]{\textcolor{gray}{#1}}
\newcommand{\best}[1]{\cellcolor{bestgreen}\textbf{#1}}
\newcommand{\second}[1]{\cellcolor{secondlime}\textbf{#1}}
\newcommand{\third}[1]{\cellcolor{thirdyellow}\textbf{#1}}

\title{\LARGE \bf
VGGT-GS SLAM: Uncalibrated Monocular Gaussian Splatting SLAM with Feed-Forward Priors
}

\author{Yuhang Han$^{1}$, Hao Wang$^{1}$, Jiaxi Cao$^{1}$ and Xingyu Liu$^{1,*}$%
\thanks{$^{1}$Authors are with the National University of Singapore, Singapore.}%
\thanks{$^{*}$Corresponding author: Xingyu Liu (e-mail: \texttt{xyl@nus.edu.sg}).}%
}

\begin{document}
\emergencystretch=2em
\hbadness=10000
\vbadness=10000
% Reclaim only external float gaps, without scaling tables or their contents.
\setlength{\floatsep}{4pt plus 2pt minus 1pt}
\setlength{\textfloatsep}{10pt plus 2pt minus 2pt}
\setlength{\dbltextfloatsep}{10pt plus 2pt minus 2pt}

\maketitle
\thispagestyle{empty}
\pagestyle{empty}

%%%%%%%%%%%%%%%%%%%%%%%%%%%%%%%%%%%%%%%%%%%%%%%%%%%%%%%%%%%%%%%%%%%%%%%%%%%%%%%%
\begin{abstract}

We present VGGT-GS SLAM, a monocular 3D Gaussian Splatting SLAM system designed for uncalibrated videos. Starting from feed-forward VGGT pose and depth priors, our system performs submap differentiable bundle adjustment that jointly refines camera poses and a 3D Gaussian map, while optimizing submap-shared intrinsics and radial--tangential distortion through analytic calibration Jacobians. To improve global consistency, we introduce Gaussian-native alignment (GNA) for camera-anchored scale refinement between sequential submaps and verification of loop-closure candidates. Extensive experiments on standard indoor benchmarks show consistent improvements in localization accuracy and strong rendering quality under uncalibrated settings, establishing a strong baseline for uncalibrated Gaussian SLAM.

\end{abstract}

%%%%%%%%%%%%%%%%%%%%%%%%%%%%%%%%%%%%%%%%%%%%%%%%%%%%%%%%%%%%%%%%%%%%%%%%%%%%%%%%
\section{INTRODUCTION}
\FloatBarrier

3D Gaussian Splatting (3DGS)~\cite{kerbl20233d} enables real-time rendering and high-fidelity novel view synthesis, motivating recent Gaussian-SLAM systems that jointly estimate camera trajectories and scene maps. However, most existing methods assume known camera calibration, and many require RGB-D input for reliable Gaussian initialization. These requirements limit deployment on commodity monocular cameras that lack calibration metadata and depth sensing. For such inputs, camera calibration must be estimated alongside motion and scene structure rather than supplied beforehand.

Recent feed-forward 3D reconstruction models, including DUSt3R~\cite{wang2024dust3r}, MASt3R~\cite{leroy2024grounding}, and VGGT~\cite{wang2025vggt}, provide strong camera pose and dense geometry priors from sets of monocular images. Systems such as GeoGS-SLAM~\cite{gao2026geogsslam} already integrate these priors with Gaussian mapping and rendering-based pose--map refinement. However, these predictions remain an initialization: errors in camera calibration can be absorbed into the estimated scene geometry, and independently reconstructed submaps need not share a consistent scale. This motivates jointly refining calibration and geometry while explicitly addressing alignment between submaps.

\begin{figure}[!t]
    \centering
    \includegraphics[width=0.95\linewidth]{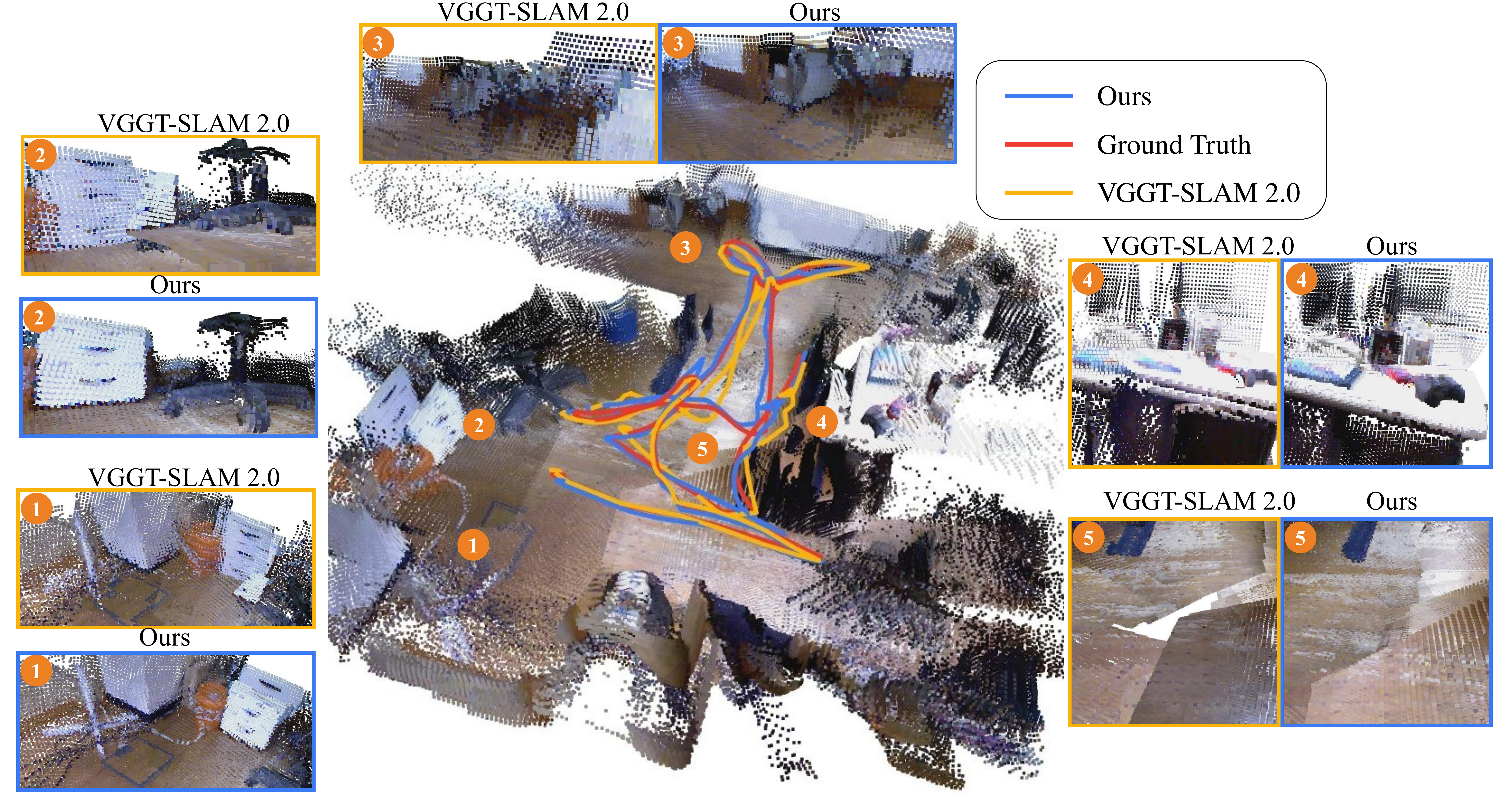}
    \caption{Qualitative trajectory and reconstruction comparison with VGGT-SLAM 2.0~\cite{maggio2026vggt} on the TUM-RGBD \textit{floor} sequence. Point clouds are visualized using input RGB pixels, VGGT-predicted depth priors, and each method's optimized poses. Our method exhibits better local geometric consistency, with less structural fragmentation and map tearing.}
    \label{fig:tum_floor_qualitative}
\end{figure}

We retain the submap organization of VGGT-SLAM~\cite{maggio2025vggt} while converting VGGT pose and depth predictions into optimizable Gaussian submaps. VGGT-predicted depth serves as a learned geometric prior rather than sensor depth input. Within each submap, we jointly refine camera poses, Gaussian attributes, and camera calibration. Analytic calibration Jacobians let image observations refine intrinsics and distortion alongside the Gaussian map. Fig.~\ref{fig:tum_floor_qualitative} provides a qualitative trajectory and reconstruction comparison.

Since submaps are optimized independently, relative scale errors can accumulate across the sequence. Our Gaussian-native matching uses both centers and learned covariances to retain spatial information beyond point locations. It refines scale alignment between consecutive submaps and verifies loop candidates before pose-graph insertion, while preserving the original transforms of accepted loops.

Our main contributions are as follows:
\begin{itemize}
\item We develop calibration-aware Gaussian bundle adjustment for uncalibrated monocular SLAM, jointly refining camera poses, Gaussians, intrinsics, and radial--tangential distortion through analytic calibration Jacobians in differentiable rendering.
\item We introduce covariance-aware Gaussian-native matching for camera-anchored sequential scale refinement and loop-candidate verification before global $\mathrm{Sim}(3)$ pose-graph optimization.
\item Extensive experiments demonstrate strong localization and rendering performance under uncalibrated monocular settings.
\end{itemize}

\section{RELATED WORK}
\label{sec:related_work}

\subsection{3D Gaussian Splatting-Based SLAM}
Early neural SLAM systems use implicit scene representations~\cite{sucar2021imap,zhu2022nice} and hybrid encodings~\cite{wang2023co} for dense tracking and mapping. In contrast, 3D Gaussian Splatting offers explicit, differentiable primitives with efficient rasterization, motivating a growing family of Gaussian-SLAM systems.

GS-SLAM develops dense RGB-D tracking and mapping with coarse-to-fine Gaussian optimization~\cite{yan2024gs}, while SplaTAM combines online Gaussian optimization with structured map expansion~\cite{keetha2024splatam}. MonoGS demonstrates a purely Gaussian-based monocular SLAM pipeline and also supports RGB-D input~\cite{matsuki2024gaussian}. Other systems combine established tracking front-ends with Gaussian mapping, including Photo-SLAM and DROID-Splat~\cite{huang2024photo,homeyer2025droid}.

To improve scalability and efficiency, Gaussian-SLAM adopts local submaps~\cite{yugay2023gaussian}.

Recent methods further address global consistency. LoopSplat registers RGB-D Gaussian submaps to construct loop constraints and optimizes a robust pose graph~\cite{zhu2025loopsplat}, while GLC-SLAM introduces hierarchical loop closure and efficient submap updates after pose correction~\cite{xu2024glc}. Our focus is covariance-aware, camera-anchored scale refinement under uncalibrated monocular input; for loops, Gaussian matching verifies proposals rather than replacing their initial transform.

\subsection{Camera Self-Calibration in Monocular SLAM}
Most visual SLAM systems assume known camera calibration. Self-calibrating approaches instead jointly estimate camera parameters, motion, and scene structure~\cite{keivan2015online}, although observability depends on camera motion and scene geometry, particularly when estimating lens distortion~\cite{zhuang2019degeneracy}.

DroidCalib embeds self-calibrating bundle adjustment into DROID-SLAM, optimizing intrinsics through differentiable Gauss--Newton steps and supporting different camera models without retraining~\cite{hagemann2023deep}. ORB-SLAM3 and DSO, by contrast, require camera calibration as input~\cite{campos2021orb,engel2017direct}.

Unlike prior methods that estimate calibration primarily from feature correspondences and geometric reprojection constraints, our method jointly refines camera intrinsics and radial--tangential distortion with camera poses and 3D Gaussians under a differentiable rendering objective within each submap.

\subsection{Feed-Forward Geometry Priors for SLAM}
DUSt3R reconstructs uncalibrated image pairs through pointmap regression~\cite{wang2024dust3r}, while MASt3R adds 3D-grounded features for correspondence matching~\cite{leroy2024grounding}. VGGT extends inference to larger image sets, jointly predicting camera parameters, depth maps, point maps, and point tracks~\cite{wang2025vggt}.

MASt3R-SLAM uses MASt3R pointmaps for tracking, local fusion, loop closure, and global optimization, assuming only a single camera center rather than a fixed parametric camera model~\cite{murai2025mast3r}. VGGT-SLAM aligns overlapping submaps on the SL(4) manifold to address projective ambiguity under unknown calibration~\cite{maggio2025vggt}. VGGT-SLAM 2.0 improves efficiency, enforces overlapping-frame calibration consistency, and introduces keyframe-level factor-graph optimization and attention-based loop verification~\cite{maggio2026vggt}.

AIM-SLAM uses information- and geometry-aware keyframe prioritization with joint multi-view Sim(3) optimization~\cite{jeon2026aimslam}. VGGT-SLAM++ combines a visual-odometry front-end with high-cadence local bundle adjustment and scalable graph construction~\cite{mandal2026vggtslampp}.

GeoGS-SLAM further combines uncalibrated feed-forward priors, Gaussian mapping, rendering-based pose--map optimization, and online loop closure~\cite{gao2026geogsslam}. Relative to this shared system design, our focus is explicit intrinsics and lens-distortion refinement through analytic rendering-chain derivatives, together with covariance-aware Gaussian matching for sequential scale refinement and loop verification. Unlike overlapping-frame calibration alignment, our local BA optimizes camera parameters against image observations jointly with the Gaussian map.

\section{METHOD}
\label{sec:method}
\begin{figure}[t]
    \centering
    \vspace*{6pt}
    \includegraphics[width=1\linewidth]{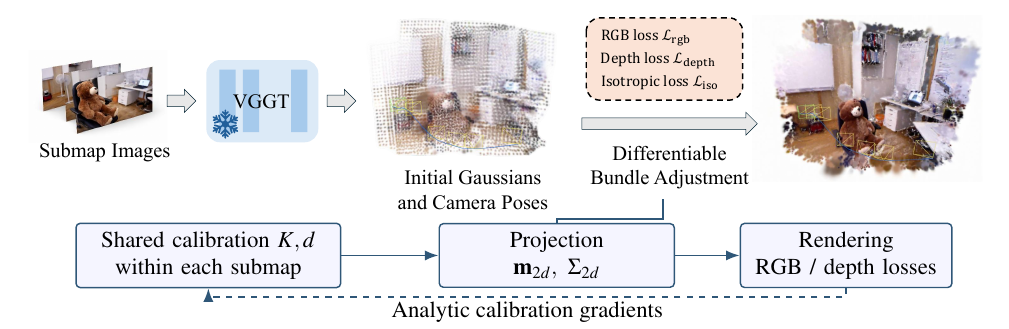}
    \caption{VGGT~\cite{wang2025vggt} pose and geometry priors initialize a Gaussian submap, whose Gaussians, camera poses, and submap-shared intrinsics $K$ and distortion $\boldsymbol d$ are jointly optimized through differentiable bundle adjustment.}
    \label{fig:dba}
\end{figure}

\FloatBarrier
We initialize Gaussian submaps from VGGT predictions and jointly refine their Gaussians, camera poses, and shared calibration through differentiable bundle adjustment (Fig.~\ref{fig:dba}). We then connect the submaps in a global pose graph using Gaussian-native $\mathrm{Sim}(3)$ constraints.

\subsection{Submaps and Initialization}
\smallskip\noindent\textbf{Submap windowing.}
Following VGGT-SLAM~\cite{maggio2025vggt}, the selected keyframes are grouped into windows of $S+O$, where $S$ is the frame number in each submap and $O$ is the overlap frame number.
\par\smallskip\noindent\textbf{VGGT initialization.}
We process each $S+O$-frame mapping window using VGGT~\cite{wang2025vggt}. VGGT predicts camera extrinsics $\{T_i^{c\leftarrow w}\}_{i=1}^{N}$, intrinsics $\{K_i\}_{i=1}^{N}$, and depth-confidence maps $\{D_i,C_i\}_{i=1}^{N}$. These predictions initialize the submap poses and geometry. We initialize the shared intrinsics $K$ by averaging $\{K_i\}$ and set the distortion coefficients to zero.

\smallskip\noindent\textbf{Gaussian initialization from depth.}
Following the initialization strategy used in MonoGS~\cite{matsuki2024gaussian}, we create an initial set of 3D Gaussians from VGGT depth priors by back-projecting valid depth pixels to 3D points in the world coordinate system using the current pose and intrinsics estimate. We subsample pixels with a fixed stride and filter invalid depths ($D_i(p) > 10^{-8}$). Each retained pixel $p$ initializes a Gaussian centered at its back-projected world point $\mu_j$, with spherical-harmonic (SH) color coefficients derived from $I_i(p)$. We use the identity rotation and set the opacity to $\alpha_j=0.5$ (zero logit). To adapt the initial support to the local sampling density, we set the isotropic log-scale to
\begin{equation}
\boldsymbol{\ell}_j=\log\!\left(0.1d_{\mathrm{nn},j}\right)\mathbf{1}_3,
\end{equation}
where $d_{\mathrm{nn},j}$ is the nearest-neighbor distance among points sampled from the same frame.

\subsection{Joint Bundle Adjustment}
\label{sec:ba}

Within each submap, we jointly optimize the Gaussian representation, camera poses, per-frame affine brightness parameters, and a camera model shared across all frames. We fix the first camera pose to define the submap reference frame and update the remaining poses using left-multiplicative $\mathrm{SE}(3)$ increments, $T_i^{c\leftarrow w}\leftarrow\exp(\xi_i)T_i^{c\leftarrow w}$ with $\xi_i\in\mathbb R^6$ for $i=2,\ldots,N$. The shared intrinsics use log-space focal-length updates and additive principal-point updates. We bound the five-parameter radial--tangential distortion coefficients through $k_j=s_k\tanh(\rho_{k_j})$ for $j=1,2,3$ and $p_j=s_p\tanh(\rho_{p_j})$ for $j=1,2$, where $s_k,s_p>0$ set their respective ranges. Analytic rasterization gradients for both intrinsics and distortion are detailed in Sec.~\ref{subsec:analytic_jacobian}.

\smallskip\noindent\textbf{Objective.}
Let $\hat I_i$, $\hat D_i$, and $\hat A_i$ denote the rendered color, expected depth, and alpha for frame $i$. We minimize
\begin{equation}
\mathcal{L}_{\mathrm{BA}}
=
\mathcal{L}_{\mathrm{rgb}}
+\lambda_{\mathrm{depth}}\mathcal{L}_{\mathrm{depth}}
+\lambda_{\mathrm{iso}}\mathcal{L}_{\mathrm{iso}}.
\end{equation}
Here $\lambda_{\mathrm{depth}}$ and $\lambda_{\mathrm{iso}}$ weight the geometric prior and isotropic regularizer, respectively. Fixing the first pose defines the local coordinate frame, whereas the VGGT depth term supplies a predicted scale reference. We minimize this objective using AdamW with parameter-specific learning rates.

\smallskip\noindent\textbf{Photometric loss.}
Following DSO~\cite{engel2017direct}, we compensate exposure changes with $I_i'(p)=\exp(a_i)I_i(p)+b_i$. On valid image pixels $\mathcal V^{I}$, excluding near-black borders, the photometric term is
\begin{equation}
\mathcal{L}_{\mathrm{rgb}}
=(1-\lambda_{\mathrm{dssim}})
\|\hat I-I'\|_{1,\mathcal V^{I}}
+\lambda_{\mathrm{dssim}}
\bigl(1-\mathrm{SSIM}(\hat I,I')\bigr).
\end{equation}
Here $\|\cdot\|_{1,\mathcal V^I}$ denotes the mean absolute error over $\mathcal V^I$, and $\lambda_{\mathrm{dssim}}\in[0,1]$ is the DSSIM mixing weight.
To address the common-darkening ambiguity between unconstrained exposure and learnable colors, an exposure-stabilized variant fixes $a_1=b_1=0$ during local BA as a brightness reference. During subsequent map refinement, this variant freezes all estimated exposure parameters while retaining exposure compensation. The main localization and NVS results use affine compensation without these additional constraints.

\smallskip\noindent\textbf{Depth loss.}
VGGT depth serves as a weak geometric prior. Define $\mathcal V_i^D=\{p\mid D_i(p)>\epsilon,\hat D_i(p)>\epsilon,\hat A_i(p)>0\}$ and the relative residual with a tolerance dead zone
\begin{equation}
r_i(p)=\left[
\frac{|\hat D_i(p)-D_i(p)|}{\max(D_i(p),\epsilon)}
-\tau_{\mathrm{depth}}
\right]_+.
\end{equation}
Here $[x]_+=\max(x,0)$, $\tau_{\mathrm{depth}}$ is the residual tolerance, and $\epsilon>0$ is a numerical stabilizer. We median-normalize and clip the VGGT confidence as $\tilde C_i(p)=\min(C_i(p)/\max(\operatorname{median}(C_i),\epsilon),c_{\max})$, where $c_{\max}$ is the confidence cap; we set $\tilde C_i(p)=1$ when confidence is unavailable. The depth term is
\begin{equation}
\mathcal{L}_{\mathrm{depth}}
=
\frac{\sum_i\sum_{p\in\mathcal V_i^D}\tilde C_i(p)r_i(p)^2}
{\sum_i\sum_{p\in\mathcal V_i^D}\tilde C_i(p)+\epsilon}.
\end{equation}

\smallskip\noindent\textbf{Isotropic regularization.}
Following MonoGS~\cite{matsuki2024gaussian}, we discourage highly anisotropic Gaussians by regularizing the physical scale $\boldsymbol{s}_j=\exp(\boldsymbol{\ell}_j)$:
\begin{equation}
\mathcal{L}_{\mathrm{iso}}
=
\frac{1}{3M}\sum_{j=1}^{M}
\left\|
\boldsymbol{s}_j-\bar{s}_j\mathbf{1}_3
\right\|_1.
\end{equation}
Here $M$ is the number of Gaussians and $\bar{s}_j$ is the mean of the three scale components.

\begin{figure*}[t]
    \centering
    \vspace*{6pt}
    \includegraphics[width=0.90\textwidth]{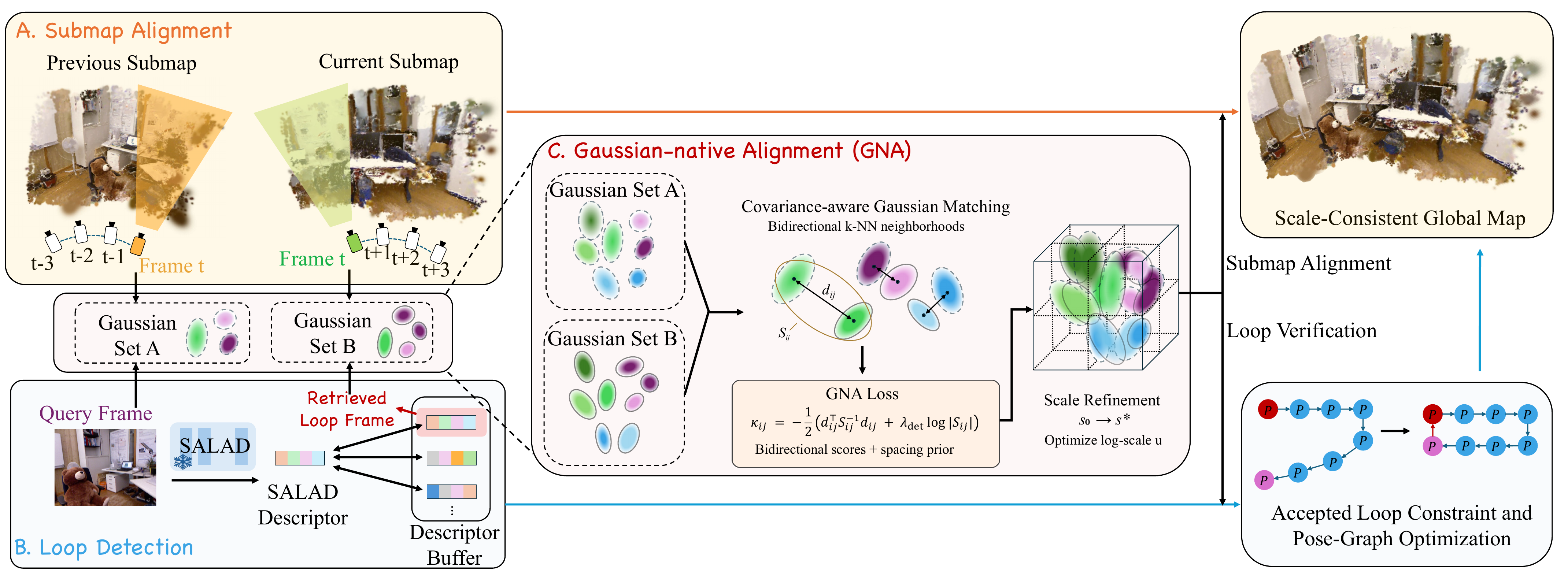}
    \caption{\textbf{Gaussian-native alignment and loop verification.} Starting from camera-tied initialization, GNA uses covariance-aware Gaussian matching to refine the relative scale between adjacent submaps. For SALAD-retrieved loop candidates, the final GNA loss and scale correction determine acceptance before pose-graph optimization; accepted loops retain the original VGGT bridge transform.}
    \label{fig:gs_align}
\end{figure*}

\subsection{Analytic Calibration Jacobians}
\label{subsec:analytic_jacobian}
We analytically differentiate the projected 2D Gaussian center $\mathbf m_{2d}$ and covariance $\Sigma_{2d}$ with respect to the full calibration vector $\boldsymbol\theta_{\mathrm{cal}}=(f_x,f_y,c_x,c_y,k_1,k_2,k_3,p_1,p_2)$ within \texttt{gsplat}'s differentiable Gaussian projection~\cite{ye2025gsplat}. These Jacobians allow the rendering loss to jointly optimize all nine calibration parameters.

\smallskip\noindent\textbf{Camera model.}
For a camera-frame point $\mathbf m=(x,y,z)^\top$, let $u=x/z$, $v=y/z$, and $r^2=u^2+v^2$. With radial factor $d=1+k_1r^2+k_2r^4+k_3r^6$, the distorted normalized coordinates $(u_d,v_d)$ and corresponding pixel coordinates $(U,V)$ are
\begin{align}
u_d &= du+2p_1uv+p_2(r^2+2u^2),\\
v_d &= dv+p_1(r^2+2v^2)+2p_2uv,\\
(U,V) &= (f_xu_d+c_x,\,f_yv_d+c_y).
\end{align}
Here $(f_x,f_y)$ are the focal lengths and $(c_x,c_y)$ is the principal point. Let $\mathbf m_d=(u_d,v_d)^\top$, $\mathbf m_{2d}=(U,V)^\top$, $\mathbf c=(c_x,c_y)^\top$, and $D=\operatorname{diag}(f_x,f_y)$, such that $\mathbf m_{2d}=D\mathbf m_d+\mathbf c$. Further, let $A=\partial(u_d,v_d)/\partial(u,v)$ and $B=\partial(u,v)/\partial(x,y,z)$.

\smallskip\noindent\textbf{Projected-center Jacobian.}
For any component $q$ of $\boldsymbol\theta_{\mathrm{cal}}$, the projected-center Jacobian is
\begin{equation}
\frac{\partial\mathbf m_{2d}}{\partial q}
=\frac{\partial D}{\partial q}\mathbf m_d
+D\frac{\partial\mathbf m_d}{\partial q}
+\frac{\partial\mathbf c}{\partial q}.
\end{equation}

\smallskip\noindent\textbf{Projected-covariance Jacobian.}
For camera-frame covariance $\Sigma_{3d}^{c}$, the projection Jacobian and image-space covariance are
\begin{equation}
J=DAB,\qquad \Sigma_{2d}=J\Sigma_{3d}^{c}J^\top.
\end{equation}
Differentiating with respect to any component $q$ of $\boldsymbol\theta_{\mathrm{cal}}$ gives
\begin{equation}
\begin{aligned}
\frac{\partial J}{\partial q}
&=\frac{\partial D}{\partial q}AB
+D\frac{\partial A}{\partial q}B,\\
\frac{\partial\Sigma_{2d}}{\partial q}
&=\frac{\partial J}{\partial q}\Sigma_{3d}^{c}J^\top
{}+J\Sigma_{3d}^{c}\left(\frac{\partial J}{\partial q}\right)^\top.
\end{aligned}
\end{equation}
The CUDA kernels return gradients with respect to the physical coefficients; automatic differentiation then applies the bounded-parameter chain rule, e.g., $\partial\mathcal L/\partial\rho_{k_j}=(\partial\mathcal L/\partial k_j)s_k(1-\tanh^2\rho_{k_j})$. These expressions describe unclipped projection.

\subsection{Gaussian-Native Alignment (GNA) for Submaps}
\label{sec:sim3_align}
Each submap has its own local frame. Adjacent submaps share an overlap keyframe. For loops, SALAD retrieves a historical frame for the current query, and joint VGGT inference initializes the inter-submap bridge (Sec.~\ref{sec:loop_closure}). Fig.~\ref{fig:gs_align} summarizes the alignment and verification paths.

\noindent\textbf{Camera-tied initialization.}
Let $A$ and $B$ denote the reference and source submaps. Their shared-image poses determine the rotation $R$ from the local frame of $B$ to that of $A$; identity-pixel depth correspondences determine the initial scale $s_0$; and the corresponding camera centers $c_A,c_B$ fix the translation as $t_0=c_A-s_0Rc_B$. We hold $R$ fixed, initialize $u=\log s_0$, and optimize only this log-scale. A source Gaussian $j$ with mean $\mu_j^B$ and spatial covariance $\Sigma_j^B$ is transformed into $A$ as
\begin{equation}
\begin{aligned}
s&=\exp(u),\\
\mu_j^{B\rightarrow A}(u)&=c_A+sR(\mu_j^B-c_B),\\
\Sigma_j^{B\rightarrow A}(u)&=s^2R\Sigma_j^B R^\top.
\end{aligned}
\end{equation}

\smallskip\noindent\textbf{Gaussian-native refinement.}
We filter Gaussians by opacity, anchor-view visibility, and spatial scale, retaining those projecting into supported pixels of the opposite anchor view. Their learned covariances encode spatial extent and orientation beyond point centers. Before evaluating overlap, we normalize the Gaussian means and covariances by a robust reference-scene scale. For a reference Gaussian $i$ and a transformed source Gaussian $j$, let $d_{ij}=\mu_i^A-\mu_j^{B\rightarrow A}$ and
$S_{ij}=\Sigma_i^A+\Sigma_j^{B\rightarrow A}+(\epsilon+\tau^2)I_3$. We use the log-kernel
\begin{equation}
\kappa_{ij}(u)=-\frac{1}{2}\left(
d_{ij}^\top S_{ij}^{-1}d_{ij}
+\lambda_{\det}\log|S_{ij}|
\right),
\end{equation}
where $I_3$ is the identity matrix, $\epsilon>0$ is a covariance stabilizer, $\tau$ is an annealed bandwidth, and $\lambda_{\det}$ controls the log-determinant term. With opacity $\alpha_i$ and observation count $n_i$, define $w_i=\max\{\alpha_i^2/[1+\log(1+n_i)],10^{-6}\}$. For the $k$ nearest reference centers $\mathcal N_A(j)$ of source center $j$, the directional score is
\begin{equation}
\ell_{B\rightarrow A}
=\frac{\sum_j w_j^B\log\!\left(
\frac{\sum_{i\in\mathcal N_A(j)}w_i^A e^{\kappa_{ij}}}
{\sum_{i\in\mathcal N_A(j)}w_i^A}\right)}{\sum_j w_j^B}.
\end{equation}
The reverse score exchanges $A$ and $B$; neighborhoods are refreshed at each bandwidth stage and fixed within it. The optimized objective is
\begin{equation}
\label{eq:gna_objective}
\mathcal L_{\mathrm{GNA}}(u)
=-\ell_{B\rightarrow A}(u)-\ell_{A\rightarrow B}(u)
+\lambda_{\mathrm{sp}}
\left(u-\log\frac{m_A}{m_B}\right)^2,
\end{equation}
where $m_A$ and $m_B$ are the median nearest-neighbor spacings of the selected Gaussian sets and $\lambda_{\mathrm{sp}}$ weights a density-sensitive heuristic scale prior. We bound the scale update around its initialization and retain it only when it improves the normalized objective; otherwise, we keep $s_0$. The resulting transform defines a sequential constraint, whereas for loop closure GNA is used only for verification as described in Sec.~\ref{sec:loop_closure}.

\subsection{Loop Closure with Gaussian-Native Verification}
\label{sec:loop_closure}
For each mapping window, we match per-frame SALAD descriptors~\cite{izquierdo2024optimal} against prior submaps, exclude the immediate predecessor, and retain at most the closest pair whose descriptor distance is below the retrieval threshold $\tau_r$. After anchor-free mapping and BA, the retrieved image is appended only to a separate joint VGGT loop-inference batch. Its predicted depth correspondences initialize the inter-submap $\mathrm{Sim}(3)$, which GNA evaluates solely for verification.

The update bounds and improvement-based fallback used for sequential alignment do not apply to loop verification; here, the scale correction is an acceptance criterion.
GNA verifies loop candidates using the final matching loss and scale correction, while accepted loops retain their original VGGT transforms:
\begin{equation}
L^\star=\mathcal L_{\mathrm{GNA}}(u^\star),\qquad
\Delta_s=\left|\log\frac{s^\star}{s_0}\right|.
\end{equation}
A loop enters the global $\mathrm{Sim}(3)$ pose graph only if GNA succeeds, $L^\star\leq\tau_L$, and $\Delta_s\leq\tau_s$; the accepted measurement remains the original VGGT bridge transform. We also do a rotation-consistency check that rejects bridges differing by more than $90^\circ$ from the preceding sequential chain.

Pose-graph nodes are local-to-global submap transforms $G_i\in\mathrm{Sim}(3)$, constrained by sequential and accepted loop measurements of $G_i^{-1}G_j$, with a strong prior anchoring the first node. Optimization updates only $G_i$, leaving local Gaussians, poses, and calibration unchanged. For $G_i=(s,R,t)$, global Gaussian means and covariances are $sR\mu+t$ and $s^2R\Sigma R^\top$; submaps are not merged.

\subsection{Other Implementation Details}
\label{sec:other_impl}
\smallskip\noindent\textbf{Keyframe selection.}
We select a new keyframe when the mean image-plane $\ell_1$ displacement of Lucas--Kanade-tracked points exceeds $\tau_{\mathrm{disp}}$, or fewer than ten valid tracks remain.

\smallskip\noindent\textbf{Gaussian refinement.}
During submap BA, we periodically clone or split Gaussians using accumulated image-space gradients and prune low-opacity ones, following MonoGS~\cite{matsuki2024gaussian}.

\smallskip\noindent\textbf{Non-keyframe pose estimation.}
After sequence-level mapping and submap $\mathrm{Sim}(3)$ graph optimization, we estimate non-keyframe poses to obtain per-frame trajectories and support novel-view synthesis. Each frame uses the local submap with the latest starting keyframe at or before it, for both pose estimation and rendering, rather than a merged global map. With fixed Gaussian maps and calibration, each pose is optimized in $\mathrm{SE}(3)$ for 90 iterations.

\section{EXPERIMENTS}

\subsection{Experimental Setup}
All experiments use RGB-only input. Experiments are performed on a server with an NVIDIA RTX PRO 6000 GPU and an AMD EPYC 9355 CPU. Unless otherwise stated, all baseline methods are locally reproduced using the authors' released implementations.

The keyframe motion threshold is $\tau_{\mathrm{disp}}=30$ pixels. Each complete mapping window contains $S=24$ new keyframes and $O=1$ overlapping keyframe. Distortion optimization is enabled by default for uncalibrated input. In the bounded distortion parameterization, $s_k=2$ and $s_p=0.05$ set the ranges of the radial and tangential coefficients, respectively. For each submap, we run 550 joint BA iterations. The depth-prior and isotropy weights are $\lambda_{\mathrm{depth}}=0.3$ and $\lambda_{\mathrm{iso}}=10$, respectively. We use a relative-depth tolerance of $\tau_{\mathrm{depth}}=0.03$ and clip the normalized VGGT confidence at $c_{\max}=5$. GNA verifier uses $\tau_L=1000$ and $\tau_s=0.20$.

Boldface indicates the best results. For uncalibrated localization and NVS results,
{\setlength{\fboxsep}{0.5pt}\colorbox{bestgreen}{\textbf{first}}},
{\setlength{\fboxsep}{0.5pt}\colorbox{secondlime}{\textbf{second}}}, and
{\setlength{\fboxsep}{0.5pt}\colorbox{thirdyellow}{\textbf{third}}}
additionally highlight the top three ranks.

\subsection{Localization Evaluation}
For localization, we evaluate on TUM-RGBD~\cite{sturm12iros}, 7-Scenes~\cite{shotton2013scene}, and ScanNet~\cite{dai2017scannet}. We compare against calibrated visual, neural, and Gaussian SLAM methods~\cite{campos2021orb,teed2018deepv2d,lipson2024dpvslam,zhang2023go,teed2021droid,murai2025mast3r,matsuki2024gaussian,zhu2024nicer,sandstrom2025splat,zhang2025hislam2}, as well as uncalibrated monocular systems~\cite{teed2021droid,murai2025mast3r,maggio2025vggt,maggio2026vggt,piedade2026slammer,hu2025ec3rslam,jeon2026aimslam,mandal2026vggtslampp,gao2026geogsslam}. Following VGGT-SLAM, uncalibrated DROID-SLAM uses GeoCalib~\cite{veicht2024geocalib} estimates. We report translational ATE RMSE in meters after $\mathrm{Sim}(3)$ alignment, using each rerun method's final native keyframes and the final graph-optimized keyframes for Ours.

\begin{table}[t]
\centering
\caption{ATE RMSE on the TUM-RGBD dataset (m).}
\label{tab:tum_ate_new}
\setlength{\tabcolsep}{3.5pt}
\renewcommand{\arraystretch}{1.02}
\resizebox{\columnwidth}{!}{%
\ADLinactivate
\begin{tabular}{@{}c|lcccccccccc}
\toprule
 & Method & 360 & desk & desk2 & floor & plant & room & rpy & teddy & xyz & Avg. \\
\midrule
\multirow{9}{*}{\rotatebox{90}{\calib{\textbf{Calib.}}}} & \calib{ORB-SLAM3~\cite{campos2021orb}} & \calib{$\times$} & \calib{\textbf{0.013}} & \calib{$\times$} & \calib{\textbf{0.015}} & \calib{0.040} & \calib{0.151} & \calib{$\times$} & \calib{0.498} & \calib{0.008} & \calib{N/A} \\
 & \calib{DeepV2D$^{\ddagger}$~\cite{teed2018deepv2d}} & \calib{0.243} & \calib{0.166} & \calib{0.379} & \calib{1.653} & \calib{0.203} & \calib{0.246} & \calib{0.105} & \calib{0.316} & \calib{0.064} & \calib{0.375} \\
 & \calib{DPV-SLAM++~\cite{lipson2024dpvslam}} & \calib{0.123} & \calib{0.019} & \calib{0.026} & \calib{0.044} & \calib{0.027} & \calib{0.382} & \calib{0.040} & \calib{0.109} & \calib{0.011} & \calib{0.087} \\
 & \calib{GO-SLAM~\cite{zhang2023go}} & \calib{0.175} & \calib{0.020} & \calib{0.038} & \calib{0.030} & \calib{0.039} & \calib{0.626} & \calib{0.025} & \calib{0.053} & \calib{0.013} & \calib{0.113} \\
 & \calib{DROID-SLAM~\cite{teed2021droid}} & \calib{0.061} & \calib{0.018} & \calib{0.028} & \calib{0.032} & \calib{\textbf{0.013}} & \calib{\textbf{0.043}} & \calib{0.030} & \calib{0.030} & \calib{0.011} & \calib{\textbf{0.030}} \\
 & \calib{MASt3R-SLAM~\cite{murai2025mast3r}} & \calib{\textbf{0.043}} & \calib{0.019} & \calib{0.026} & \calib{0.029} & \calib{0.018} & \calib{0.073} & \calib{0.030} & \calib{0.038} & \calib{\textbf{0.006}} & \calib{0.031} \\
 & \calib{MonoGS~\cite{matsuki2024gaussian}} & \calib{0.175} & \calib{0.564} & \calib{0.704} & \calib{0.392} & \calib{0.090} & \calib{0.900} & \calib{0.049} & \calib{0.113} & \calib{0.016} & \calib{0.334} \\
 & \calib{Splat-SLAM~\cite{sandstrom2025splat}} & \calib{0.056} & \calib{0.019} & \calib{0.028} & \calib{0.030} & \calib{0.018} & \calib{0.044} & \calib{\textbf{0.022}} & \calib{0.042} & \calib{0.012} & \calib{\textbf{0.030}} \\
 & \calib{HI-SLAM2~\cite{zhang2025hislam2}} & \calib{0.155} & \calib{0.016} & \calib{\textbf{0.022}} & \calib{0.035} & \calib{0.015} & \calib{0.057} & \calib{0.025} & \calib{\textbf{0.027}} & \calib{0.010} & \calib{0.040} \\
\midrule
\multirow{10}{*}{\rotatebox{90}{\textbf{Uncalib.}}} & DROID-SLAM~\cite{teed2021droid} & 0.189 & 0.035 & 0.796 & 0.278 & 0.034 & 0.829 & 0.054 & 0.038 & \third{0.013} & 0.252 \\
 & MASt3R-SLAM~\cite{murai2025mast3r} & 0.072 & 0.038 & 0.059 & \third{0.050} & 0.038 & 0.098 & 0.041 & 0.135 & 0.020 & 0.061 \\
 & VGGT-SLAM~\cite{maggio2025vggt} & 0.068 & 0.027 & 0.042 & 0.157 & \second{0.023} & 0.117 & 0.032 & 0.039 & 0.014 & 0.058 \\
 & VGGT-SLAM 2.0~\cite{maggio2026vggt} & \second{0.050} & 0.026 & 0.029 & 0.100 & \third{0.025} & 0.064 & \third{0.026} & \third{0.035} & 0.014 & 0.041 \\
 & SLAM-MER~\cite{piedade2026slammer} & 0.103 & 0.034 & 0.039 & 0.221 & 0.044 & 0.228 & 0.050 & 0.118 & \best{0.006} & 0.094 \\
 & EC3R-SLAM~\cite{hu2025ec3rslam} & 0.107 & 0.070 & 0.081 & 0.054 & 0.088 & 0.136 & 0.046 & 0.112 & 0.030 & 0.080 \\
 & AIM-SLAM$^{\ddagger}$~\cite{jeon2026aimslam} & \second{0.050} & \best{0.017} & \third{0.028} & \best{0.024} & 0.026 & \third{0.062} & \best{0.021} & 0.039 & \second{0.010} & \second{0.031} \\
 & VGGT-SLAM++$^{\ddagger}$~\cite{mandal2026vggtslampp} & \best{0.042} & \third{0.025} & \second{0.027} & 0.077 & 0.042 & \best{0.027} & \third{0.026} & \best{0.029} & 0.016 & 0.036 \\
 & GeoGS-SLAM$^{\ddagger}$~\cite{gao2026geogsslam} & -- & -- & -- & -- & -- & -- & -- & -- & -- & \third{0.035} \\
 & Ours & \third{0.058} & \second{0.021} & \best{0.026} & \second{0.029} & \best{0.020} & \second{0.050} & \second{0.024} & \second{0.033} & \third{0.013} & \best{0.030} \\
\bottomrule
\end{tabular}}
\vspace{1pt}

\begin{minipage}{\columnwidth}
\fontsize{6}{7}\selectfont\raggedright
$\ddagger$: Paper-reported results. GeoGS-SLAM reports only the average over the same nine sequences; its paper does not provide per-sequence values.\\
VGGT-SLAM++'s reported Avg.\ of 0.036 is retained; its listed per-sequence values average 0.0346.\\
$\times$: Less than 70\% time-span coverage.
\end{minipage}
\end{table}

As shown in Table~\ref{tab:tum_ate_new}, our method achieves the lowest average ATE among uncalibrated systems on TUM-RGBD, obtaining 0.0304\,m compared with 0.031\,m for AIM-SLAM~\cite{jeon2026aimslam}. Ours ranks first on desk2 and plant and second on five additional sequences, with comparable accuracy to calibrated Splat-SLAM~\cite{sandstrom2025splat} (0.030\,m). On 7-Scenes sequence 01 (Table~\ref{tab:7scenes_ate_new}), ours achieves the best uncalibrated average of 0.055\,m, improving VGGT-SLAM++~\cite{mandal2026vggtslampp} (0.064\,m) by 14.1\%. It ranks first on office and redkitchen and second on chess, fire, and stairs. Finally, on the six ScanNet sequences in Table~\ref{tab:scannet_ate_new}, ours ranks first among uncalibrated methods on five sequences, reducing the average ATE from 0.103\,m for VGGT-SLAM~2.0 to 0.090\,m.

\begin{table}[t]
\centering
\caption{ATE RMSE on 7-Scenes sequence 01 (m).}
\label{tab:7scenes_ate_new}
\setlength{\tabcolsep}{3.5pt}
\renewcommand{\arraystretch}{1.02}
\resizebox{0.927\columnwidth}{!}{%
\ADLinactivate
\begin{tabular}{@{}c|lcccccccc}
\toprule
 & Method & chess & fire & heads & office & pumpkin & redkitchen & stairs & Avg. \\
\midrule
\multirow{9}{*}{\rotatebox{90}{\calib{\textbf{Calib.}}}} & \calib{ORB-SLAM3~\cite{campos2021orb}} & \calib{0.043} & \calib{0.030} & \calib{0.069} & \calib{0.092} & \calib{0.116} & \calib{0.046} & \calib{0.272} & \calib{0.095} \\
 & \calib{DPV-SLAM++~\cite{lipson2024dpvslam}} & \calib{0.038} & \calib{0.044} & \calib{0.027} & \calib{0.064} & \calib{0.120} & \calib{\textbf{0.036}} & \calib{0.025} & \calib{0.051} \\
 & \calib{GO-SLAM~\cite{zhang2023go}} & \calib{0.040} & \calib{0.033} & \calib{0.027} & \calib{\textbf{0.055}} & \calib{0.118} & \calib{0.038} & \calib{0.023} & \calib{0.048} \\
 & \calib{NICER-SLAM$^{\ddagger}$~\cite{zhu2024nicer}} & \calib{0.033} & \calib{0.069} & \calib{0.042} & \calib{0.108} & \calib{0.200} & \calib{0.039} & \calib{0.108} & \calib{0.086} \\
 & \calib{DROID-SLAM~\cite{teed2021droid}} & \calib{\textbf{0.032}} & \calib{0.028} & \calib{0.025} & \calib{0.080} & \calib{0.088} & \calib{0.041} & \calib{0.017} & \calib{\textbf{0.045}} \\
 & \calib{MASt3R-SLAM~\cite{murai2025mast3r}} & \calib{0.059} & \calib{\textbf{0.025}} & \calib{\textbf{0.014}} & \calib{0.096} & \calib{\textbf{0.081}} & \calib{0.042} & \calib{\textbf{0.010}} & \calib{0.047} \\
 & \calib{MonoGS~\cite{matsuki2024gaussian}} & \calib{0.043} & \calib{0.056} & \calib{0.445} & \calib{0.479} & \calib{0.234} & \calib{0.063} & \calib{0.210} & \calib{0.218} \\
 & \calib{Splat-SLAM~\cite{sandstrom2025splat}} & \calib{0.038} & \calib{0.032} & \calib{0.029} & \calib{\textbf{0.055}} & \calib{0.113} & \calib{0.042} & \calib{0.020} & \calib{0.047} \\
 & \calib{HI-SLAM2~\cite{zhang2025hislam2}} & \calib{0.036} & \calib{0.032} & \calib{0.026} & \calib{0.080} & \calib{0.116} & \calib{0.040} & \calib{0.027} & \calib{0.051} \\
\midrule
\multirow{8}{*}{\rotatebox{90}{\textbf{Uncalib.}}} & DROID-SLAM~\cite{teed2021droid} & \third{0.038} & 0.037 & 0.307 & 0.166 & \third{0.132} & 0.090 & \best{0.018} & 0.113 \\
 & MASt3R-SLAM~\cite{murai2025mast3r} & 0.063 & 0.046 & 0.029 & 0.113 & \best{0.116} & 0.075 & 0.064 & 0.072 \\
 & VGGT-SLAM~\cite{maggio2025vggt} & \second{0.037} & \second{0.025} & \second{0.018} & \third{0.098} & 0.139 & 0.071 & 0.092 & 0.069 \\
 & VGGT-SLAM 2.0~\cite{maggio2026vggt} & \third{0.038} & \third{0.026} & \second{0.018} & 0.102 & 0.134 & \third{0.060} & 0.092 & \third{0.067} \\
 & SLAM-MER~\cite{piedade2026slammer} & 0.052 & 0.028 & 0.060 & \second{0.097} & 0.153 & \second{0.058} & \third{0.039} & 0.070 \\
 & EC3R-SLAM~\cite{hu2025ec3rslam} & 0.086 & 0.045 & 0.055 & 0.126 & 0.171 & 0.067 & 0.041 & 0.084 \\
 & VGGT-SLAM++$^{\ddagger}$~\cite{mandal2026vggtslampp} & \best{0.034} & \best{0.023} & \best{0.017} & 0.104 & \second{0.127} & 0.085 & 0.060 & \second{0.064} \\
 & Ours & \second{0.037} & \second{0.025} & \third{0.020} & \best{0.096} & 0.136 & \best{0.039} & \second{0.034} & \best{0.055} \\
\bottomrule
\end{tabular}}
\vspace{1pt}

\begin{minipage}{\columnwidth}
\fontsize{6}{7}\selectfont\raggedright
\begin{tabular}{@{}l@{}}
$\ddagger$: Paper-reported results.
\end{tabular}
\end{minipage}
\end{table}

\begin{table}[t]
\centering
\caption{ATE RMSE on the ScanNet dataset (m).}
\label{tab:scannet_ate_new}
\setlength{\tabcolsep}{3.5pt}
\renewcommand{\arraystretch}{1.02}
\resizebox{0.760\columnwidth}{!}{%
\ADLinactivate
\begin{tabular}{@{}c|lccccccc}
\toprule
 & Method & 0000 & 0059 & 0106 & 0169 & 0181 & 0207 & Avg. \\
\midrule
\multirow{8}{*}{\rotatebox{90}{\calib{\textbf{Calib.}}}} & \calib{ORB-SLAM3~\cite{campos2021orb}} & \calib{0.071} & \calib{0.118} & \calib{$\times$} & \calib{\textbf{0.073}} & \calib{0.179} & \calib{0.559} & \calib{N/A} \\
 & \calib{DPV-SLAM++~\cite{lipson2024dpvslam}} & \calib{0.086} & \calib{0.214} & \calib{0.126} & \calib{0.091} & \calib{0.117} & \calib{0.301} & \calib{0.156} \\
 & \calib{GO-SLAM~\cite{zhang2023go}} & \calib{0.055} & \calib{0.097} & \calib{0.152} & \calib{0.099} & \calib{0.085} & \calib{0.076} & \calib{0.094} \\
 & \calib{DROID-SLAM~\cite{teed2021droid}} & \calib{0.055} & \calib{0.815} & \calib{0.830} & \calib{0.092} & \calib{0.117} & \calib{0.433} & \calib{0.390} \\
 & \calib{MASt3R-SLAM~\cite{murai2025mast3r}} & \calib{0.060} & \calib{0.082} & \calib{0.086} & \calib{0.084} & \calib{\textbf{0.070}} & \calib{\textbf{0.066}} & \calib{\textbf{0.075}} \\
 & \calib{MonoGS~\cite{matsuki2024gaussian}} & \calib{1.547} & \calib{0.893} & \calib{1.501} & \calib{1.996} & \calib{0.910} & \calib{0.900} & \calib{1.291} \\
 & \calib{Splat-SLAM~\cite{sandstrom2025splat}} & \calib{0.062} & \calib{0.092} & \calib{\textbf{0.082}} & \calib{0.088} & \calib{0.079} & \calib{0.070} & \calib{0.079} \\
 & \calib{HI-SLAM2~\cite{zhang2025hislam2}} & \calib{\textbf{0.054}} & \calib{\textbf{0.078}} & \calib{1.300} & \calib{0.084} & \calib{0.072} & \calib{\textbf{0.066}} & \calib{0.275} \\
\midrule
\multirow{7}{*}{\rotatebox{90}{\textbf{Uncalib.}}} & DROID-SLAM~\cite{teed2021droid} & 1.655 & 0.728 & 0.153 & \second{0.107} & 0.449 & 0.443 & 0.589 \\
 & MASt3R-SLAM~\cite{murai2025mast3r} & \third{0.106} & \third{0.097} & \third{0.113} & \third{0.136} & 0.110 & \third{0.123} & \third{0.114} \\
 & VGGT-SLAM~\cite{maggio2025vggt} & 0.194 & 0.100 & 0.175 & 0.181 & 0.148 & 0.126 & 0.154 \\
 & VGGT-SLAM 2.0~\cite{maggio2026vggt} & 0.116 & \second{0.092} & \second{0.097} & 0.142 & \second{0.080} & \second{0.090} & \second{0.103} \\
 & SLAM-MER~\cite{piedade2026slammer} & 0.211 & 0.106 & 0.228 & 0.144 & 0.211 & 0.166 & 0.178 \\
 & EC3R-SLAM~\cite{hu2025ec3rslam} & \best{0.104} & 0.117 & 0.171 & 0.238 & \third{0.094} & 0.284 & 0.168 \\
 & Ours & \second{0.105} & \best{0.086} & \best{0.092} & \best{0.096} & \best{0.079} & \best{0.083} & \best{0.090} \\
\bottomrule
\end{tabular}}
\vspace{1pt}

\begin{minipage}{\columnwidth}
\fontsize{6}{7}\selectfont\raggedright
\begin{tabular}{@{}l@{}}
$\times$: Less than 70\% time-span coverage.
\end{tabular}
\end{minipage}
\end{table}

\begin{figure*}[!t]
    \centering
    \vspace*{6pt}
    \includegraphics[width=0.95\textwidth]{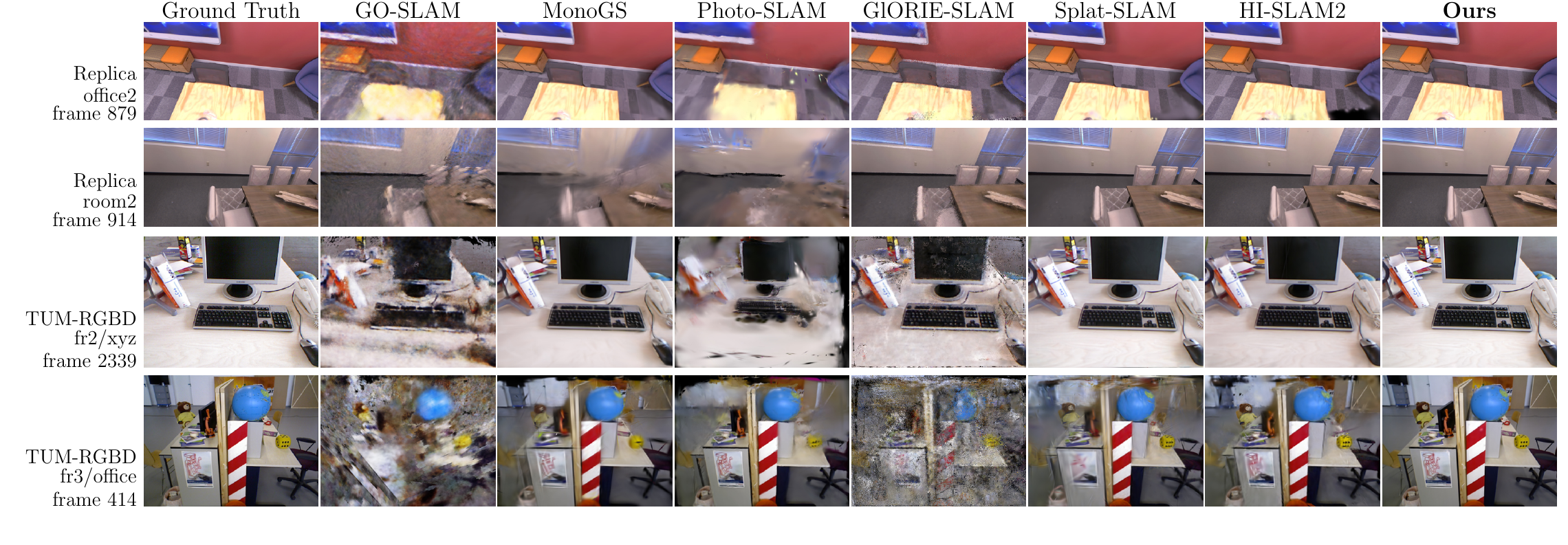}
    \caption{Qualitative NVS examples from archived Replica and TUM-RGBD runs, with ours in the rightmost column.}
    \label{fig:nvs_qualitative}
\end{figure*}

\subsection{Novel View Synthesis}
We evaluate NVS on Replica~\cite{straub2019replica} and TUM-RGBD~\cite{sturm12iros} against RGB-only neural and Gaussian SLAM methods~\cite{zhang2023go,matsuki2024gaussian,huang2024photo,zhang2024glorie,sandstrom2025splat,zhang2025hislam2}, using each work's official implementation. We evaluate every fifth frame, excluding mapping keyframes and using a common frame set across methods. Replica and TUM use $640\times360$ and $512\times384$ images, respectively; calibrated methods use undistorted RGB, while ours uses raw RGB. For TUM scoring, we bilinearly remap our saved renders to the common undistorted domain using dataset calibration only for evaluation. All methods share the same reference RGB and geometrically valid crop, with all RGB channels scored for PSNR. We score saved 8-bit RGB outputs using PSNR, SSIM, and AlexNet-based LPIPS~\cite{zhang2018unreasonable}, averaging frames within scenes and then scenes equally.

For NVS, TUM uses the BA maps directly; Replica adds 500 mapping-refinement and 500 RGB-only refinement iterations per submap, with poses and calibration fixed.

Our method achieves the best average PSNR and LPIPS on both datasets
(Tables~\ref{tab:replica_nvs_metrics} and~\ref{tab:tum_nvs_metrics}).
On Replica, it achieves 39.100\,dB PSNR and 0.020 LPIPS, compared with
HI-SLAM2's 39.046\,dB and 0.029. On TUM, our PSNR is comparable to Splat-SLAM (24.484 vs.\ 24.413\,dB), with lower LPIPS than HI-SLAM2 (0.185 vs.\ 0.197), which achieves the best SSIM. Bicubic remapping preserves these mean rankings. Fig.~\ref{fig:nvs_qualitative} presents qualitative comparisons.

% Tables: nvs_pixel_domain_tum_results.json (TUM, common_linear),
% nvs_corrected_main_tables.json (Replica baselines),
% replica_budget_500_500_results.json (all Replica adapted/common Ours cells).
% Means use unrounded scene scores; highlights use displayed three-decimal ties.
% Archived qualitative figure is not a rendering of the newly selected table settings.

\begin{table}[t]
\centering
\caption{Replica NVS on common frames.}
\label{tab:replica_nvs_metrics}
\fontsize{6}{6.6}\selectfont
\setlength{\tabcolsep}{0.6pt}
\renewcommand{\arraystretch}{0.90}
\setlength{\aboverulesep}{0.24ex}
\setlength{\belowrulesep}{0.24ex}
\begin{tabular}{llccccccccc}
\arrayrulecolor{black}
\toprule
Method & Metric & room0 & room1 & room2 & office0 & office1 & office2 & office3 & office4 & Avg. \\
\midrule
\arrayrulecolor{black!25}
\multirow{3}{*}{GO-SLAM~\cite{zhang2023go}}
 & PSNR $\uparrow$ & 21.794 & 25.527 & 24.001 & 28.349 & 30.088 & 22.281 & 23.138 & 24.730 & 24.989 \\
 & SSIM $\uparrow$ & 0.625 & 0.759 & 0.732 & 0.819 & 0.868 & 0.742 & 0.725 & 0.805 & 0.759 \\
 & LPIPS $\downarrow$ & 0.570 & 0.449 & 0.483 & 0.430 & 0.319 & 0.454 & 0.449 & 0.455 & 0.451 \\
\midrule
\multirow{3}{*}{MonoGS~\cite{matsuki2024gaussian}}
 & PSNR $\uparrow$ & 29.618 & 28.638 & 24.023 & 32.291 & 35.970 & 27.429 & 29.915 & 28.882 & 29.596 \\
 & SSIM $\uparrow$ & 0.896 & 0.849 & 0.804 & 0.903 & 0.944 & \third{0.898} & 0.925 & 0.923 & 0.893 \\
 & LPIPS $\downarrow$ & 0.095 & 0.161 & 0.333 & 0.158 & 0.096 & 0.169 & 0.092 & 0.143 & 0.156 \\
\midrule
\multirow{3}{*}{Photo-SLAM~\cite{huang2024photo}}
 & PSNR $\uparrow$ & 25.065 & 18.801 & 20.301 & 23.441 & 23.886 & 18.360 & 25.087 & 22.940 & 22.235 \\
 & SSIM $\uparrow$ & 0.785 & 0.710 & 0.759 & 0.768 & 0.778 & 0.770 & 0.856 & 0.837 & 0.783 \\
 & LPIPS $\downarrow$ & 0.238 & 0.570 & 0.486 & 0.466 & 0.390 & 0.379 & 0.175 & 0.325 & 0.379 \\
\midrule
\multirow{3}{*}{GlORIE-SLAM~\cite{zhang2024glorie}}
 & PSNR $\uparrow$ & 27.261 & 28.394 & 28.664 & 33.848 & 17.123 & 26.531 & 27.098 & 11.568 & 25.061 \\
 & SSIM $\uparrow$ & 0.851 & 0.872 & 0.885 & 0.928 & 0.427 & 0.874 & 0.877 & 0.568 & 0.785 \\
 & LPIPS $\downarrow$ & 0.179 & 0.184 & 0.190 & 0.112 & 0.692 & 0.212 & 0.160 & 0.753 & 0.310 \\
\midrule
\multirow{3}{*}{Splat-SLAM~\cite{sandstrom2025splat}}
 & PSNR $\uparrow$ & \third{33.273} & \third{35.472} & \third{36.297} & \third{41.053} & \third{40.809} & \third{35.701} & \third{36.284} & \third{38.288} & \third{37.147} \\
 & SSIM $\uparrow$ & \third{0.930} & \third{0.940} & \third{0.960} & \third{0.976} & \third{0.971} & \second{0.946} & \third{0.963} & \third{0.970} & \third{0.957} \\
 & LPIPS $\downarrow$ & \third{0.064} & \third{0.067} & \third{0.052} & \third{0.040} & \third{0.053} & \third{0.097} & \third{0.043} & \third{0.048} & \third{0.058} \\
\midrule
\multirow{3}{*}{HI-SLAM2~\cite{zhang2025hislam2}}
 & PSNR $\uparrow$ & \best{35.888} & \best{37.233} & \best{38.503} & \second{42.440} & \best{43.089} & \second{37.746} & \best{38.117} & \second{39.355} & \second{39.046} \\
 & SSIM $\uparrow$ & \best{0.957} & \best{0.970} & \best{0.976} & \second{0.983} & \best{0.981} & \best{0.975} & \best{0.975} & \second{0.978} & \best{0.975} \\
 & LPIPS $\downarrow$ & \second{0.039} & \second{0.033} & \second{0.027} & \second{0.021} & \best{0.022} & \second{0.032} & \second{0.027} & \second{0.030} & \second{0.029} \\
\midrule
\multirow{3}{*}{Ours}
 & PSNR $\uparrow$ & \second{35.398} & \second{36.871} & \second{37.669} & \best{43.334} & \second{42.185} & \best{38.700} & \second{38.017} & \best{40.623} & \best{39.100} \\
 & SSIM $\uparrow$ & \second{0.956} & \second{0.962} & \second{0.969} & \best{0.986} & \second{0.978} & \best{0.975} & \second{0.973} & \best{0.982} & \second{0.973} \\
 & LPIPS $\downarrow$ & \best{0.026} & \best{0.019} & \best{0.019} & \best{0.012} & \second{0.024} & \best{0.022} & \best{0.022} & \best{0.018} & \best{0.020} \\
\arrayrulecolor{black}
\bottomrule
\arrayrulecolor{black}
\end{tabular}
\end{table}

\begin{table}[t]
\centering
\caption{TUM-RGBD NVS on common frames.}
\label{tab:tum_nvs_metrics}
\fontsize{6}{6.6}\selectfont
\setlength{\tabcolsep}{1.4pt}
\renewcommand{\arraystretch}{0.90}
\setlength{\aboverulesep}{0.24ex}
\setlength{\belowrulesep}{0.24ex}
\begin{tabular}{llcccccc}
\arrayrulecolor{black}
\toprule
Method & Metric & f1/desk & f2/xyz & f3/off & f1/desk2 & f1/room & Avg. \\
\midrule
\arrayrulecolor{black!25}
\multirow{3}{*}{GO-SLAM~\cite{zhang2023go}}
 & PSNR $\uparrow$ & 15.137 & 14.218 & 14.024 & 5.626 & 15.065 & 12.814 \\
 & SSIM $\uparrow$ & 0.489 & 0.462 & 0.433 & 0.077 & 0.495 & 0.391 \\
 & LPIPS $\downarrow$ & 0.587 & 0.571 & 0.714 & 0.861 & 0.634 & 0.674 \\
\midrule
\multirow{3}{*}{MonoGS~\cite{matsuki2024gaussian}}
 & PSNR $\uparrow$ & 16.241 & 24.243 & 22.858 & 15.934 & 15.067 & 18.868 \\
 & SSIM $\uparrow$ & 0.587 & 0.826 & 0.788 & 0.579 & 0.536 & 0.663 \\
 & LPIPS $\downarrow$ & 0.503 & 0.154 & 0.267 & 0.510 & 0.543 & 0.395 \\
\midrule
\multirow{3}{*}{Photo-SLAM~\cite{huang2024photo}}
 & PSNR $\uparrow$ & 14.072 & 15.572 & 17.795 & 12.942 & 12.288 & 14.534 \\
 & SSIM $\uparrow$ & 0.507 & 0.563 & 0.641 & 0.486 & 0.457 & 0.531 \\
 & LPIPS $\downarrow$ & 0.528 & 0.625 & 0.371 & 0.596 & 0.663 & 0.557 \\
\midrule
\multirow{3}{*}{GlORIE-SLAM~\cite{zhang2024glorie}}
 & PSNR $\uparrow$ & 14.195 & 19.063 & 16.906 & 13.095 & 13.868 & 15.426 \\
 & SSIM $\uparrow$ & 0.315 & 0.622 & 0.403 & 0.249 & 0.279 & 0.374 \\
 & LPIPS $\downarrow$ & 0.839 & 0.453 & 0.738 & 0.903 & 0.903 & 0.767 \\
\midrule
\multirow{3}{*}{Splat-SLAM~\cite{sandstrom2025splat}}
 & PSNR $\uparrow$ & \best{25.497} & \second{26.922} & \third{23.078} & \second{22.617} & \best{23.953} & \second{24.413} \\
 & SSIM $\uparrow$ & \second{0.837} & \third{0.857} & \third{0.791} & \third{0.748} & \best{0.780} & \third{0.803} \\
 & LPIPS $\downarrow$ & \third{0.177} & \second{0.122} & \third{0.239} & \third{0.283} & \best{0.244} & \third{0.213} \\
\midrule
\multirow{3}{*}{HI-SLAM2~\cite{zhang2025hislam2}}
 & PSNR $\uparrow$ & \second{24.568} & \best{27.563} & \second{23.549} & \third{22.611} & \third{21.459} & \third{23.950} \\
 & SSIM $\uparrow$ & \best{0.849} & \best{0.900} & \second{0.828} & \best{0.825} & \second{0.761} & \best{0.833} \\
 & LPIPS $\downarrow$ & \best{0.173} & \best{0.083} & \second{0.210} & \second{0.213} & \third{0.304} & \second{0.197} \\
\midrule
\multirow{3}{*}{Ours}
 & PSNR $\uparrow$ & \third{24.459} & \third{26.567} & \best{24.847} & \best{24.144} & \second{22.403} & \best{24.484} \\
 & SSIM $\uparrow$ & \third{0.831} & \second{0.864} & \best{0.856} & \second{0.814} & \third{0.758} & \second{0.825} \\
 & LPIPS $\downarrow$ & \second{0.175} & \third{0.123} & \best{0.163} & \best{0.208} & \second{0.254} & \best{0.185} \\
\arrayrulecolor{black}
\bottomrule
\arrayrulecolor{black}
\end{tabular}
\end{table}

\subsection{More Experiments}
\setcounter{topnumber}{4}
Unless otherwise specified, all experiments in this section use all Freiburg1 sequences from TUM-RGBD~\cite{sturm12iros}.

% Detailed protocols/results: paper/review/calibration_revision_results.md,
% calibration_full_camera_metrics.json, overlap_consistency_results.md,
% calibration_jacobian_fix_results.md. Calibration diagnostics use a revised
% configuration and predate the FoV-gradient fix; gradient tests/benchmarks
% use the corrected kernels. Neither replaces historical main-table runs.
\smallskip\noindent\textbf{Calibration validation.}
Our CUDA calibration gradients match float64 PyTorch autograd references within $4.1\times10^{-7}$ relative $L_2$ error, including reparameterization and FoV clipping. For $10^3$--$10^5$ Gaussians, FP32 projection forward/backward is $6.1$--$12.0\times$ faster than the matched PyTorch reference, including map and camera gradients, as measured using warmed CUDA graphs.

Under the default configuration, calibration refinement across 95 Freiburg1 submaps reduces sequence-averaged projection error from 14.791 to 11.185\,px and ray error from $1.435^\circ$ to $1.059^\circ$, evaluated at $41\times31$ uniformly spaced sample points across the $640\times480$ image. Intrinsics are expressed at $640\times480$ resolution. Projection error measures the pixel displacement obtained by projecting reference-camera rays through the estimated camera model; ray error measures the angle between their back-projected rays at the same pixel. In a six-window control, refining intrinsics reduces projection error from 13.982 to 10.627\,px; jointly refining intrinsics and distortion achieves a similar error of 10.746\,px.

\smallskip\noindent\textbf{Ablation study.}
Table~\ref{tab:ablation_summary} evaluates each component by removing it individually. Removing the depth loss also removes the depth-based scale reference, without adding another scale anchor. GNA has the largest impact: removing it increases ATE RMSE from 0.0304 to 0.0720\,m. Disabling calibration optimization raises the error to 0.0324\,m, while removing exposure compensation, the depth loss, or the isotropic regularizer also reduces localization accuracy.

\begin{table}[t]
\centering
\caption{Component ablation on TUM-RGBD.}
\label{tab:ablation_summary}
\scriptsize
\setlength{\tabcolsep}{3pt}
\renewcommand{\arraystretch}{0.85}
\newcommand{\cmark}{\textcolor{green!60!black}{\ensuremath{\boldsymbol{\checkmark}}}}
\newcommand{\xmark}{\textcolor{red!75!black}{\ensuremath{\boldsymbol{\times}}}}
\begin{tabular}{cccccc}
\toprule
GNA & Calib. opt. & Affine & $\mathcal{L}_{\mathrm{depth}}$ & $\mathcal{L}_{\mathrm{iso}}$ & ATE RMSE $\downarrow$ \\
\midrule
\xmark & \cmark & \cmark & \cmark & \cmark & 0.0720 \\
\cmark & \xmark & \cmark & \cmark & \cmark & 0.0324 \\
\cmark & \cmark & \xmark & \cmark & \cmark & 0.0313 \\
\cmark & \cmark & \cmark & \xmark & \cmark & 0.0316 \\
\cmark & \cmark & \cmark & \cmark & \xmark & 0.0311 \\
\cmark & \cmark & \cmark & \cmark & \cmark & \textbf{0.0304} \\
\bottomrule
\end{tabular}
\end{table}

\smallskip\noindent\textbf{GNA matching ablation.}
We evaluate all 86 consecutive submap pairs across the nine TUM fr1 sequences. Table~\ref{tab:gna_mechanism} compares Gaussian matching objectives with loop closures disabled and identical Gaussian selections and optimization settings. Matching Gaussian distributions instead of their centers reduces mean ATE by 11.8\%. Accounting for Gaussian shape and orientation provides a small improvement over spherical Gaussians, while the spacing prior has little effect in this experiment. In a separate density test on nine submap pairs, uniformly retaining 25\% of the Gaussians on either side changes the estimated relative scale by at most 0.80\% compared with full density, under the same update bounds and acceptance criteria.

% Sources: gna_mechanism_results.md/.json, gna_density_optimization_results.md/.json.
% Fixed revised-calibration maps, separate from the main-table runs.
% Full edge-scale diagnostics remain in the review records; the table focuses on ATE.
% Edge-scale references use GT-aligned local trajectories, not exact surface scale.
\begin{table}[t]
\centering
\caption{GNA matching ablation. Sequence-averaged ATE with loop closures disabled.}
\label{tab:gna_mechanism}
\scriptsize
\setlength{\tabcolsep}{4pt}
\begin{tabular}{lc}
\toprule
Alignment variant & ATE (m) $\downarrow$ \\
\midrule
Initial alignment (no GNA) & 0.0583 \\
Center-only matching & 0.0604 \\
GNA with spherical Gaussians & 0.0539 \\
GNA without spacing prior & 0.0534 \\
\textbf{Full GNA (Ours)} & \textbf{0.0533} \\
\bottomrule
\end{tabular}
\end{table}

\smallskip\noindent\textbf{GNA loop-closure verification.}
On TUM Freiburg1, SALAD~\cite{izquierdo2024optimal} retrieval ($\tau_r=1.05$, top-5; top-1 online) yields 357 proposals, of which 347 have valid ground-truth associations. A 15\% bidirectional GT-depth overlap threshold identifies 255 valid loops and 92 false matches. Table~\ref{tab:loop_verification} shows that GNA without the rotation check retains more valid loops (249 vs.\ 247) with fewer false accepts (3 vs.\ 7) than the official VGGT-SLAM~2.0 verifier~\cite{maggio2026vggt} at its code threshold of 0.95, with or without the same check. The check accepts 83 false matches, but removes one additional false match when combined with GNA, without losing valid loops.

In a separate fixed-map replay with online top-1 retrieval, loop verification drives most localization gains; adding GNA alignment between consecutive submaps further reduces mean ATE by 2.6\%. Together, they reduce mean 1-s rotational RPE RMSE from $3.181^\circ$ to $1.520^\circ$. Rotational RPE measures the error in relative rotation between keyframes approximately 1\,s apart ($\pm0.1$\,s, without interpolation). For identical accepted loops, the original transforms outperform GNA-refined ones, supporting verification rather than transform refinement for loop closures.
% Updated replay: paper/review/unified_gna_b_replay_results.json (B verifier).
% Raw/refined bridge comparison: gna_development_claims.json; separate fixed-acceptance control.

% Sources: gna_verifier_repair_results.json, gna_development_claims.json,
% gna_prospective_conclusions.md, gna_prospective_protocol.md,
% and gna_paper_integration_plan.md.
% New pool/maps, not a replay of the old 215 proposals or historical main table.
% Historical objective + 90-degree guard is an opt-in verifier variant.
% Missing-GT candidates remain scored and in graph replay; classification excludes them.
% Rotation-only and shared-guard controls are reported below; threshold curves,
% edge-quality and less favorable additional-scene recall remain in review records.

\begin{table}[t]
    \centering
    \caption{Offline loop verification.}
    \label{tab:loop_verification}
    \scriptsize
    \setlength{\tabcolsep}{2.5pt}
    \renewcommand{\arraystretch}{0.85}
    \begin{tabular}{lrrrrcc}
        \toprule
        Verifier & TP & FN & FP & TN &
        Pos. recall $\uparrow$ & Neg. rej. $\uparrow$ \\
        \midrule
        Rotation check only
            & \textbf{255} & \textbf{0} & 83 & 9 & \textbf{100.00\%} & 9.78\% \\
        VGGT-SLAM~2.0~\cite{maggio2026vggt}
            & 247 & 8 & 7 & 85 & 96.86\% & 92.39\% \\
        \quad + rotation check
            & 247 & 8 & 7 & 85 & 96.86\% & 92.39\% \\
        GNA w/o rotation check
            & 249 & 6 & 3 & 89 & 97.65\% & 96.74\% \\
        \textbf{Full GNA (Ours)}
            & 249 & 6 & \textbf{2} & \textbf{90} &
              97.65\% & \textbf{97.83\%} \\
        \bottomrule
    \end{tabular}
\end{table}

\smallskip\noindent\textbf{Submap-size trade-off.}
Table~\ref{tab:submap_tradeoff} reports performance under different submap size averaged over nine Freiburg1. Both FPS metrics divide the number of input frames by processing time: Backend FPS covers the SLAM stage only while Full FPS additionally includes loading, non-keyframe pose estimation. The 24-keyframe setting (default) gives the lowest average ATE (0.030\,m) in the accuracy study.

% Runtime/memory: paper/review/submap_size_runtime_results.md/.json; validated v3 runs.
% ATE: restored previous paper table, not the single-sequence v3 ATE results.
% All six use 550 BA steps, wd=0.01, skip_final_adaptation=true, gap=0,
% and one overlap frame. The runtime-comparison TUM Ours row also uses s24.
% Mean memory is full-process time-weighted; peaks are approximately 1-s samples.

\begin{table}[t]
\centering
\caption{Submap-size trade-off, GPU memory is in GiB.}
\label{tab:submap_tradeoff}
\scriptsize
\setlength{\tabcolsep}{3pt}
\renewcommand{\arraystretch}{0.85}
\begin{tabular}{cccccc}
\toprule
size & ATE (m) $\downarrow$ & Full FPS $\uparrow$ & Backend FPS $\uparrow$ & Avg. GPU $\downarrow$ & Peak GPU $\downarrow$ \\
\midrule
2  & 0.040 & 1.453 & 2.044 & 7.211 & \textbf{10.277} \\
4  & 0.039 & 2.086 & 3.611 & 6.609 & 11.730 \\
8  & 0.040 & 2.709 & 6.141 & 5.619 & 12.510 \\
16 & 0.037 & 3.045 & 9.653 & 4.071 & 12.729 \\
\textbf{24} & \textbf{0.030} & 3.157 & 10.765 & 3.745 & 13.281 \\
32 & 0.035 & \textbf{3.376} & \textbf{13.086} & \textbf{3.362} & 13.512 \\
\bottomrule
\end{tabular}
\end{table}

\smallskip\noindent\textbf{Runtime and GPU memory comparison.}
Table~\ref{tab:runtime_memory_comparison} compares runtime and peak GPU memory on TUM fr1/xyz and the first 1,000 frames of Replica office0. The table shows baselines slower than Ours on both sequences. Timing includes loading, SLAM, non-keyframe pose estimation without additional refinement. Ours achieves 3.157/3.29 FPS on TUM/Replica, respectively, running $1.58/1.47\times$ faster than MonoGS and $2.01/1.75\times$ faster than Splat-SLAM. Non-keyframe pose estimation accounts for 69.0\% and 57.1\% of total runtime on TUM and Replica, respectively.

% TUM Ours: submap_size_runtime_results.json, validated v3/s24,
% process_wall_seconds=252.80476592900231, tracking_seconds=174.3403485910967.
% Replica and baseline timing sources are unchanged; historical runs remain archived.

\begin{table}[t]
\centering
\caption{Runtime and sampled peak GPU memory (GiB).}
\label{tab:runtime_memory_comparison}
\scriptsize
\setlength{\tabcolsep}{3pt}
\renewcommand{\arraystretch}{0.85}
\begin{tabular}{lrrrr}
\toprule
& \multicolumn{2}{c}{TUM fr1/xyz} & \multicolumn{2}{c}{Replica office0} \\
\cmidrule(lr){2-3}\cmidrule(lr){4-5}
Method & Time (s) $\downarrow$ & Peak $\downarrow$ & Time (s) $\downarrow$ & Peak $\downarrow$ \\
\midrule
MonoGS~\cite{matsuki2024gaussian} & 399.61 & \textbf{2.40} & 447.85 & \textbf{2.90} \\
Splat-SLAM~\cite{sandstrom2025splat} & 508.48 & 11.35 & 531.45 & 11.65 \\
GlORIE-SLAM~\cite{zhang2024glorie} & 1814.17 & 15.93 & 3672.51 & 15.36 \\
\textbf{Ours} & \textbf{252.80} & 13.28 & \textbf{304.23} & 14.24 \\
\bottomrule
\end{tabular}
\end{table}

\begin{table}[t]
\centering
\caption{Generalization across feed-forward frontends.}
\label{tab:frontend_generalization}
\scriptsize
\setlength{\tabcolsep}{6pt}
\renewcommand{\arraystretch}{0.85}
\begin{tabular}{lccc}
\toprule
Frontend & Raw $\downarrow$ & +Backend $\downarrow$ & Improvement \\
\midrule
DA3-Large~\cite{lin2025depthanything3} & 0.064 & 0.033 & 49.0\% \\
Fast3R~\cite{yang2025fast3r} & 0.178 & 0.137 & 22.7\% \\
VGGT~\cite{wang2025vggt} & 0.074 & \textbf{0.0304} & \textbf{58.9\%} \\
\bottomrule
\end{tabular}
\end{table}

\smallskip\noindent\textbf{Feed-forward frontend generalization.}
To test whether the proposed backend depends specifically on VGGT~\cite{wang2025vggt}, we replace the feed-forward prior with DA3-Large~\cite{lin2025depthanything3} and Fast3R~\cite{yang2025fast3r} while keeping the 24-keyframe submap setting fixed. Raw includes submap alignment, pose-graph optimization, and loop closure, rather than isolated network predictions. Table~\ref{tab:frontend_generalization} reports ATE RMSE in meters: the same Gaussian BA and GNA backend improves all three frontends, reducing average ATE by 22.7--58.9\%.

\section{CONCLUSION}
We present VGGT-GS SLAM, a monocular dense SLAM framework that jointly estimates camera poses, Gaussian maps, intrinsics, and lens distortion from uncalibrated RGB input. By combining feed-forward geometric priors with calibration-aware Gaussian bundle adjustment and Gaussian-native submap alignment and loop verification, our method achieves strong localization and novel-view synthesis performance on indoor benchmarks. Its performance remains sensitive to prior quality and calibration observability, while local map deformation and window-induced latency remain limitations.

\FloatBarrier
\bibliographystyle{IEEEtran}
\bibliography{main}

\end{document}